\newif\ifarxiv
\newif\ifralfinal
\newif\ifconffinal
\ralfinalfalse \arxivtrue \conffinalfalse

\ifarxiv
\documentclass[letterpaper, 10 pt, conference]{ieeeconf}  %
\fi
\ifralfinal
\documentclass[letterpaper, 10 pt, journal, twoside]{IEEEtran}
\fi
\ifconffinal
\documentclass[letterpaper, 10 pt, conference]{ieeeconf} %
\fi

\IEEEoverridecommandlockouts

\usepackage{graphics} %
\usepackage{epsfig} %
\usepackage{times} %
\usepackage{amsmath} %
\usepackage{amssymb}  %
\usepackage{bbm}
\usepackage{booktabs}
\usepackage{mathtools}
\usepackage{multirow}
\usepackage{units}
\usepackage{bm}
\usepackage{xcolor}
\let\labelindent\relax
\usepackage{enumitem}
\usepackage[ruled,vlined]{algorithm2e}
\usepackage{algorithmicx}
\usepackage{cite} 
\usepackage[caption=false, font=footnotesize]{subfig}
\usepackage{graphicx}
\usepackage{adjustbox}
\usepackage{letltxmacro}
\LetLtxMacro{\originaleqref}{\eqref}
\renewcommand{\eqref}{Eq.~\originaleqref}

\SetCommentSty{mycommfont} %
\usepackage{microtype}

\ifarxiv
\usepackage{fancyhdr}
\fi
\makeatletter
\let\NAT@parse\undefined
\makeatother
\usepackage{hyperref}  %

\begin{document}
\title{
\ifarxiv\LARGE \bf\fi
DisPlacing Objects: Improving Dynamic Vehicle Detection via Visual Place Recognition under Adverse Conditions
}
\author{
\ifarxiv
Stephen Hausler\authorrefmark{2}, Sourav Garg\authorrefmark{2}, Punarjay Chakravarty\authorrefmark{4}, \\ Shubham Shrivastava\authorrefmark{3}, Ankit Vora\authorrefmark{3} and Michael Milford$^*$\authorrefmark{2}%
\fi
\thanks{\ifarxiv\authorrefmark{2}\fi 
The authors are with the QUT Centre for Robotics, School of Electrical Engineering and Robotics at the Queensland University of Technology.}%
\thanks{$^*$E-mail: \emph{firstname}.\emph{lastname}@qut.edu.au.}%
\thanks{\ifarxiv\authorrefmark{3}\fi
The authors are with the Ford Motor Company.}%
\thanks{\ifarxiv\authorrefmark{4}\fi
This work was conducted while the author was at the Ford Motor Company.}
\ifarxiv
\thanks{This research has been supported by the Ford-QUT Alliance, NVIDIA, the QUT Centre for Robotics and ARC Laureate Fellowship FL210100156.}%
\fi
}
\bstctlcite{IEEEexample:BSTcontrol}

\maketitle
\ifarxiv
\thispagestyle{fancy}
\pagestyle{plain}
\fi

\begin{abstract}

Can knowing where you are assist in perceiving objects in your surroundings, especially under adverse weather and lighting conditions? In this work we investigate whether a prior map can be leveraged to aid in the detection of dynamic objects in a scene without the need for a 3D map or pixel-level map-query correspondences. We contribute an algorithm which refines an initial set of candidate object detections and produces a refined subset of highly accurate detections using a prior map. We begin by using visual place recognition (VPR) to retrieve a reference map image for a given query image, then use a binary classification neural network that compares the query and mapping image regions to validate the query detection. Once our classification network is trained, on approximately 1000 query-map image pairs, it is able to improve the performance of vehicle detection when combined with an existing off-the-shelf vehicle detector.
We demonstrate our approach using standard datasets across two cities (Oxford and Zurich) under different settings of train-test separation of map-query traverse pairs. We further emphasize the performance gains of our approach against alternative design choices and show that VPR suffices for the task, eliminating the need for precise ground truth localization.

\end{abstract}
\section{Introduction}

Perception is a key component of autonomous vehicles, necessary in order to detect and track vehicles and pedestrians in the environment around the vehicle for its safe operation. This can be a very challenging task, especially when the driving conditions deteriorate and visibility is reduced such as at night-time~\cite{Maddern2017,lengyel2021zero}. %

Localization is another key requirement for autonomous vehicles. Localization can be divided into coarse localization using GPS or Visual Place Recognition (VPR)~\cite{lowry2016, Arandjelovic16, hausler2021patch} and fine-grained 6-DoF metric localization using techniques based on 3D reconstruction~\cite{schoenberger2016sfm} and local feature matching~\cite{Sarlin19}. Furthermore, a duality exists between perception and localization: knowing where you are can aid perception and vice-versa.

\begin{figure}
    \centering
    \includegraphics[width=\textwidth, trim=0cm 2.2cm 0cm 0cm,clip]{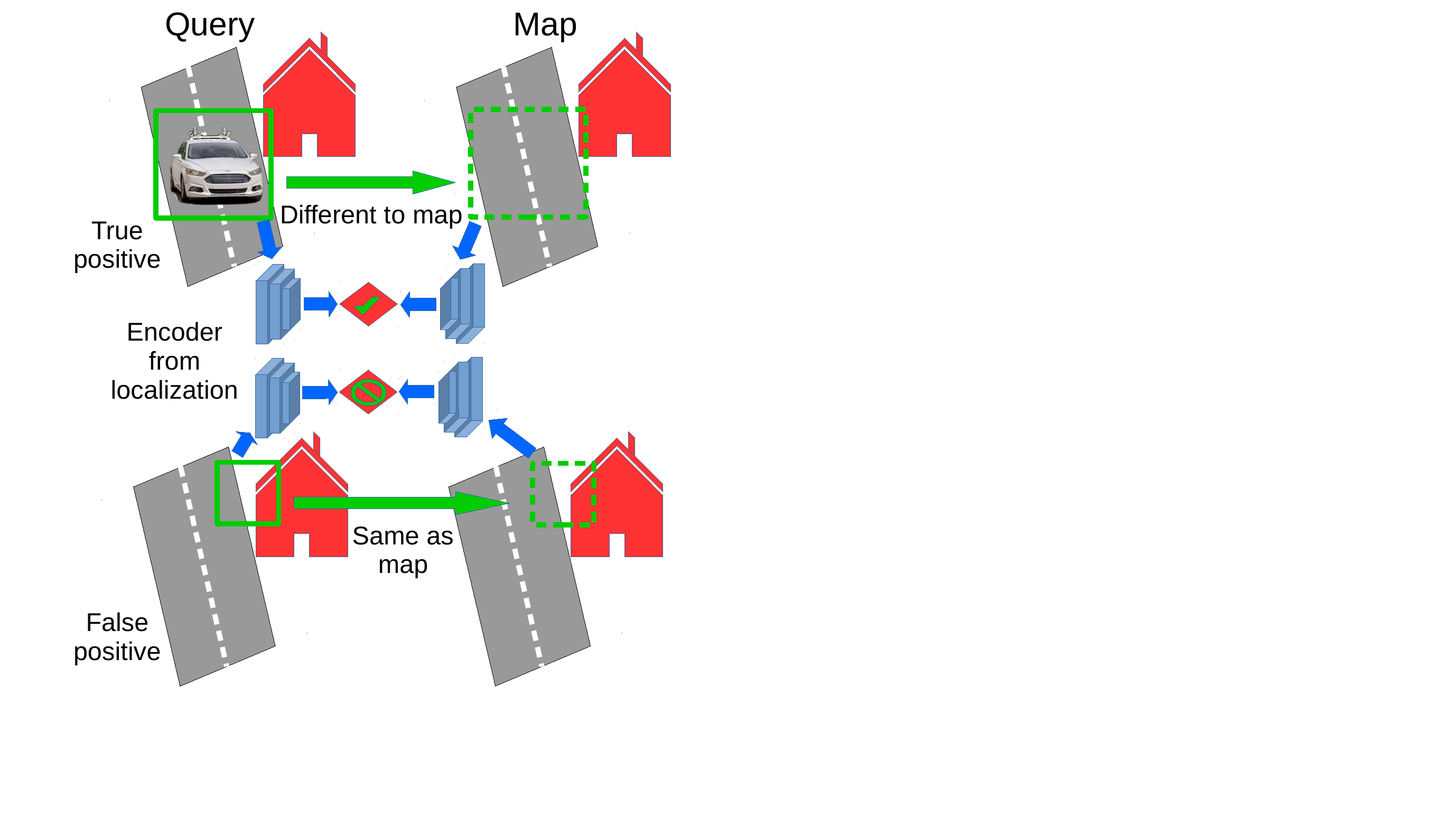}
    \caption{The aim of this work is to distinguish between true and false positive vehicle detections based on the similarity between the currently observed scene and a prior map, assuming successful localization through visual place recognition.}
    \label{fig:frontpage}
\end{figure}

In this work, we consider whether localization can aid in the detection of potentially dynamic objects such as vehicles, particularly for safe AV operations under adverse weather and lighting conditions. How can knowing where you are improve perception? Localization grants access to a prior map of a given location, which could be as simple as a reference database image. This can then be used to estimate which parts of the currently observed scene are the same as the prior map -- a dynamic vehicle will appear as a deviation from the prior map. However, a caveat exists: the current scene may change due to other effects, such as structural changes to the environment or a different viewpoint of the scene than that experienced during mapping. Furthermore, this map comparison or change detection process also needs to be invariant to changes induced by weather and lighting effects. In this paper, we approach the problem of dynamic vehicle detection from an appearance-invariant, repeated-route place recognition perspective, posed as a problem of change detection between the query and its recognized reference image. 

Our proposed approach uses an existing off-the-shelf 2D vehicle detection method to detect a large initial set of candidate vehicles. We then improve the precision of this high-recall detector through a lightweight MLP (Multi-Layer Perceptron) that compares the query image detection regions with the corresponding regions in the reference map image, obtained through VPR. Our comparator function trained only on the Oxford Robotcar dataset~\cite{Maddern2017} is tested on three different settings of map-query traverse pairs: revisiting the train route, a route geographically disparate from the train route but in the same city (Oxford), and finally, in a different city (Zurich)~\cite{SDV20}.

\section{Related works}

\subsection{Visual Place Recognition and Localization}
Both visual place recognition and visual localization are highly active areas of research, with recent respective benchmarks~\cite{toft2020long,berton2022deep} demonstrating high recognition performance. Local feature matching~\cite{hausler2021patch,DeTone18,sarlin2020superglue} is key to such performance, typically preceded by compact global descriptor based fast retrieval of candidate matches~\cite{Arandjelovic16,cao2020unifying,berton2022rethinking}. While the output of a 6-DoF localization system is a 3D pose, the aim of VPR is to find the most similar-appearing image from the reference database (map). Since we focus on 2D object detection from images in this work, we use Patch-NetVLAD as our VPR technique to retrieve a map image, which is then used to improve object detection in the query image. 

\subsection{2D Object Detection}
2D Object detection from monocular images is a well-studied problem with recent advances heavily relying on deep learning~\cite{zou2023object,liu2020deep,garg2020semantics}, including popular earlier methods like Faster RCNN~\cite{girshick2015fast} and YOLO~\cite{redmon2016you}. Some of the existing works focus on detecting 3D bounding boxes, where the input can be 3D point clouds~\cite{yang2018pixor,pan20213d} or a 2D image alone~\cite{chen2016monocular,xu2018multi}. More recent work has focused on improving object detection under adverse weather conditions such as night, rain, snow and fog. This includes adversarial training of domain-invariant features~\cite{chen2018domain}, learning weather-specific priors~\cite{sindagi2020prior}, multi-scale feature learning per domain~\cite{hnewa2021multiscale}, image-level feature alignment for single stage detectors~\cite{zhang2021domain}, image enhancement before object detection~\cite{hu2018exposure,huang2020dsnet} which includes approaches specific to hazy~\cite{li2017aod,liu2019griddehazenet,dong2020multi} and low light conditions~\cite{he2010single,lengyel2021zero} as well as using multiple differentiable image pre-processing units in sequence~\cite{liu2022image} or parallel~\cite{kalwar2022gdip}. These research efforts have been further fostered by the availability of annotated datasets for adverse conditions (e.g., BDD100K~\cite{Yu2020}), leading to robust supervised methods such as YOLOP~\cite{Wu2022}. Despite training on adverse conditions, this problem still remains challenging. In the autonomous driving scenario, where a vehicle repeatedly traverses the same route with similar viewpoints of places but under different weather conditions, it is possible to use the prior map to enhance object detection when revisiting that place, especially for detecting dynamic vehicles to avoid collisions.

\subsection{Prior Maps for Perception}
Using prior maps to improve perception during deployment has been explored before, as demonstrated on depth estimation~\cite{patil2022improving}, semantic segmentation~\cite{sakaridis2020map}, 3D object detection~\cite{fujimoto2022lanefusion,carrillo2021urbannet,ravi2018real}, visual odometry~\cite{Driven2Distract}, motion planning~\cite{chen2020learning,zeng2019end}, ego-lane detection~\cite{wang2020map} and change detection for road infrastructure either using 3D maps~\cite{lambert2trust,pannen2019hd,heo2020hd} or through image-to-image comparison~\cite{wang2014cdnet,alcantarilla2018street}. Despite the rapidly growing literature, existing approaches have limitations: they either rely on 3D maps (with significant variations in what map elements are used)~\cite{patil2022improving,fujimoto2022lanefusion,carrillo2021urbannet,ravi2018real,You2022}; only focus on updating static parts of the map~\cite{wang2020map,lambert2trust,pannen2019hd,heo2020hd}; require information from multiple prior traverses at test time~\cite{You2022}; or require pixel-level correspondences when using image-to-image setting~\cite{sakaridis2020map,Driven2Distract}
In this work, for the task of dynamic vehicle detection in 2D, we consider a repeated route scenario for an autonomous vehicle. Thus, without needing a 3D map or pixel-level correspondences (especially under adverse conditions), we show how dynamic vehicle detections in the query image can be validated by comparing corresponding regions in the query and the reference image obtained through VPR.

\section{Methodology}

In this work, our objective is to enhance the ability of an autonomous vehicle to detect vehicles in the surrounding environment, even in visually challenging situations such as at night-time. We propose that, by knowing where an autonomous vehicle is located, we can leverage prior maps of the environment to better detect dynamic vehicles in the environment. 
More specifically, our approach can be summarized as follows. Given a candidate region in the query image, produced by an off-the-shelf object detector, we propose to compare it with its corresponding region in the reference map image to determine whether this region appears the same, or different, to the prior map. If this candidate detection is a false detection, then the visual appearance of this region is likely to be the same as that in the prior map.

However, naively comparing the pixel data between a query and a mapping image is problematic as the visual appearance between these images may change due to appearance changes induced by weather, seasonal or lighting conditions. Therefore, we propose to utilize the deep features of a VPR method (which is used to localize the autonomous vehicle) as an appearance-invariant representation of image regions. We then contribute a lightweight MLP as a map matching neural network, which classifies whether a given detection candidate is correct or not, based on the similarity between the query and the map regions.

\subsection{Localization Component}

We use visual place recognition (VPR) to coarsely localize the autonomous vehicle in the prior map. Our approach can use any existing deep-learnt VPR technique, e.g., global descriptor methods such as NetVLAD~\cite{Arandjelovic16} or CosPlace~\cite{berton2022rethinking}, or local descriptor methods such as DELG~\cite{cao2020unifying} or Patch-NetVLAD~\cite{hausler2021patch}. For all our experimental results, we use Patch-NetVLAD for retrieving the reference map image for a given query image. In our classifier network (described later), we reuse the backbone features of our VPR method, which
avoids the requirement of a separate encoder and saves compute time.

\subsection{Object Detection}

We propose to use an off-the-shelf object detection method to extract a large set of potential candidate vehicles in the images observed by the autonomous vehicle. In our experiments, we use the recent vehicle detection network YOLOP~\cite{Wu2022}. YOLOP is a variant of YOLO that has been trained on BDD100K~\cite{Yu2020} to perform the tasks of vehicle detection, lane line segmentation and driveable area segmentation. Because YOLOP has been trained solely to detect vehicles, it is already a high-performing baseline detector. Additionally, BDD100K contains data collected during adverse environmental conditions, which allows YOLOP to operate well even under challenging conditions such as night-time and snow.

We modify YOLOP by reducing its default detection threshold -- this is to ensure that the recall of the detector is as high as possible, since any missed detections at this stage will not be recoverable. Our core premise is to use this high-recall detector to improve its detection precision by removing false positives using our proposed map matching classifier.

\begin{figure*}[t!]  %
    \centering
    \includegraphics[width=0.8\textwidth, trim=0cm 3.5cm 2.9cm 0cm,clip]{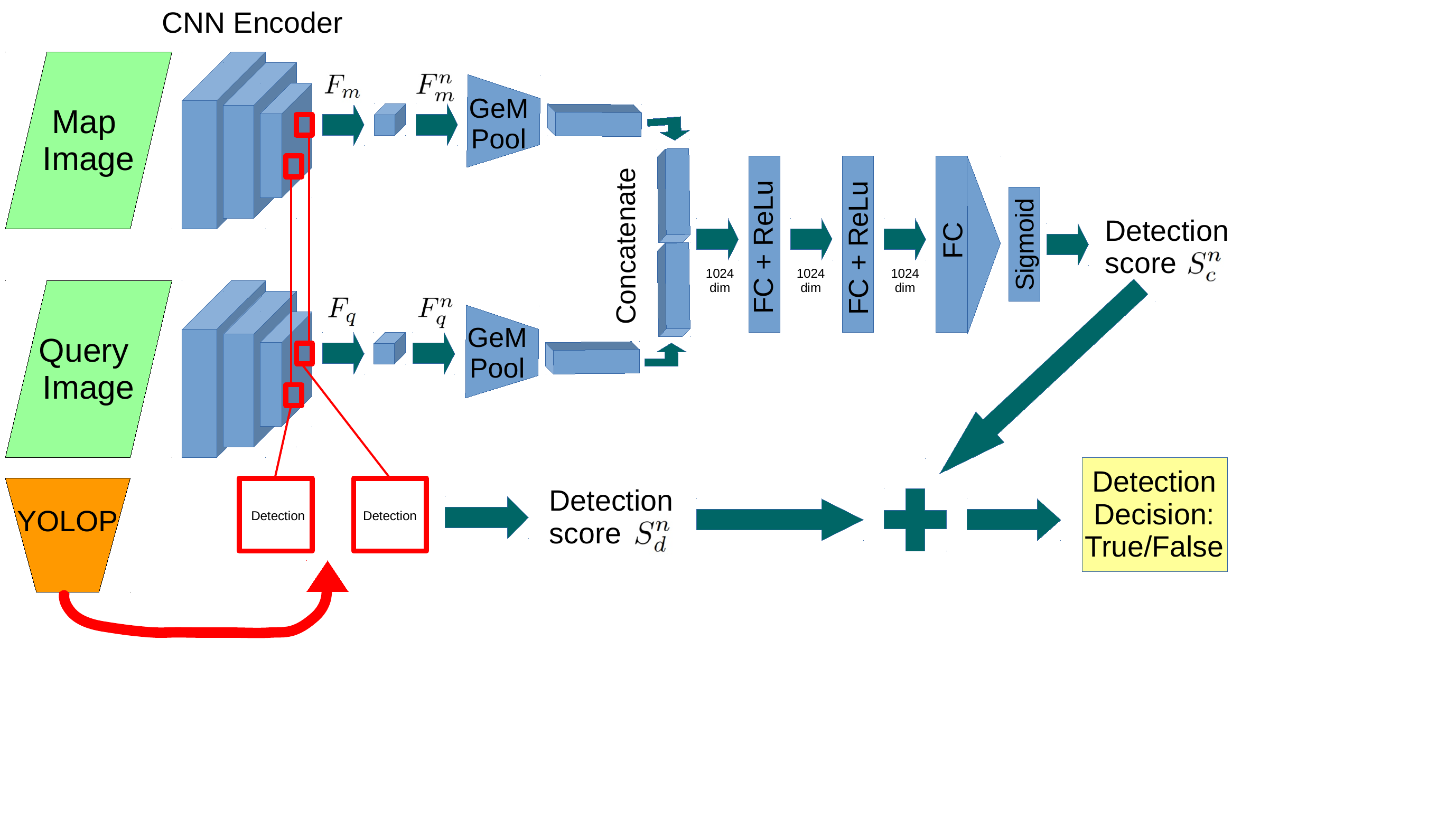}
    \caption{System diagram of our proposed approach.}
    \label{fig:arch}
\end{figure*}

\subsection{Map Matching Classifier}

Given an individual detection from YOLOP and whole-image deep features $F_q$ (query image) and $F_m$ (reference map image) from our VPR neural network, we want to decide whether this detection is correct or not. We begin by resizing the detection bounding box (in pixel coordinates) to fit the (smaller) scale size of the feature map $F_q$. We then extract a subset of $F_q$, denoted as $F_q^n$, where $n$ is the detection index in the current query image; this subset is the area of $F_q$ that overlaps with the resized detection bounding box. We then extract a subset of $F_m$ at the same position as $F_q^n$, denoted as $F_m^n$. We repeat this process for all detections $n \in N$.

In our experiments, we leverage the fact that images are collected at a high-frame rate for AV operations, thus for a repeated-route similar-viewpoint scenario, we can assume that a given downsized bounding box can be replicated at the same pixel location on both $F_m$ and $F_q$.
We assume here that localization is successful, since incorrect localization will cause this system to fail (see Table~\ref{tab:gt} for a corresponding ablation study), although we note that additional modalities such as GPS can be used to improve the reliability of the localization component.

We propose to use a classification network to decide whether $F_q^n$ and $F_m^n$ are the same or a dynamic vehicle is present in the query image. It is also worth noting that vehicles located in the mapping images can potentially show up as map differences, however, a neural network can learn to emphasise vehicles located in the query images.

\subsubsection{Classification Network Architecture}

Our map matching network consists of two heads, taking in both $F_q^n$ and $F_m^n$. We GeM pool~\cite{GeM} both $F_q^n$ and $F_m^n$ to produce individual $C$ dimensional vectors, where $C$ is the number of channels of the final conv layer of the VPR network. We concatenate both vectors together to produce the input encoding $E$. We then use an MLP as a three layer fully connected network with intermediate ReLU activations and dropout, and a sigmoid activation layer at the end (see Figure~\ref{fig:arch}). The sigmoid activation ensures the output range between 0 and 1. The network is trained using binary cross entropy loss:
\begin{equation}
    BCE = y_n \log x_n+(1-y_n)\log(1-x_n),
\end{equation}
where $y_n$ is the ground truth for a given detection candidate $n$, and $x_n$ is the predicted likelihood for a detection, where 1 denotes a true detection.

\subsubsection{Final Decision}

At this point, we could just use the output of our classification network to decide whether a detection candidate is correct or not. However, that approach would discard the detection confidence scores produced by our underlying object detector. Through experimentation, we observed that combining the confidence scores of both the map comparison classification network and from YOLOP led to more accurate vehicle detections than either confidence score in isolation. Therefore, for the final detection decision, we average the output of the classification network and the initial YOLOP detection score together:
\begin{equation}
    S^n = 0.5(S_c^n + S_d^n)
\end{equation}
where $S_c^n$ is the value from the classification network and $S_d^n$ is the detection confidence from YOLOP for the initial detection $n$.
We can then make a refined detection decision on the basis of the value of $S^n$, thresholded by $\tau$. This process is repeated for each detection $n \in N$, where $N$ is the initial larger set of noisy detection candidates produced by YOLOP.

\section{Experimental Setup}
Here, we present details of the two public benchmark datasets that we used, our object annotation procedure for Oxford, evaluation metrics and other implementation details.

\subsection{Datasets}

\subsubsection{Oxford RobotCar}

The RobotCar dataset~\cite{Maddern2017} consists of a collection of vehicle traverses through Oxford, recorded over multiple times of day and across seasons. From this dataset, a standard 6-DoF localization benchmark exists, called RobotCar Seasons v2 (RCSv2)~\cite{toft2020long}. The RCSv2 dataset contains multiple traverses of the same driving route, with images collected from the left, right and rear cameras on the vehicles. The dataset contains a mapping traverse, which was collected during overcast daytime conditions during winter (in 2014); we continue to use this traverse for all our map images. We then use the different query conditions in the training and test sets of RCSv2, with our own custom train, validation and test splits.

Our training conditions are sun, overcast summer, snow, dawn and night-rain (all rear camera). In total, our training set contains 1122 image pairs, that is one image from a query condition and the second from the mapping set (overcast-reference). Our validation condition is dusk, using the rear camera. All images for training and validation are from the `training' set of RCSv2.

Our test traverses are collected from both the training and test sets of RCSv2, for the conditions overcast-winter, rain, night and dusk and using different cameras. It is important to note that any query images from the training set use the same mapping images as the neural network was trained on. In our results, these traverses include the suffix \emph{same map}. This setting allows us to evaluate a `map-specific repeated route' operation of an autonomous vehicle. To show complete separation between training and test sets, we experiment using query images from the test set of RCSv2, where the mapping images are unseen by the classification neural network. These traverses include the suffix \emph{geosep}. Since RCSv2 does not include the front camera, we have also added traverse \emph{2015/11/13} from the original RobotCar dataset, which is the same timestamp as overcast-winter in RCSv2.

\subsubsection{Dark Zurich}

The Dark Zurich dataset~\cite{SDV20} is a driving dataset collected around Zurich, across three different times of day; denoted as daytime, twilight and nighttime. All three times of day use the same driving route. We use the daytime traverse as our mapping set, and use the nighttime traverse as our query set, where our objective is to improve the baseline detection performance of YOLOP. We \textit{do not} fine-tune our network on Dark Zurich and use the network that was trained on RCSv2. The Dark Zurich also contains pixel-wise semantic segmentation annotations, which we only use to evaluate our vehicle detection performance.

\subsection{Annotation Procedure to Generate the Ground Truth}

Since the RobotCar Seasons dataset does not provide ground truth object detections or semantic segmentation, we produced our own manually-annotated vehicle detections for query images in the RobotCar Seasons dataset. In this work, we only consider whether a given detection is correct or not, we do not analyse whether the bounding box shape is exactly the right size. Our objective is to decide whether a given detection bounding box contains the observable centroid of the true vehicle in the image. Our annotation consists of a one pixel binary flag, where each observed vehicle in an image is given a single flag located at the centroid of each vehicle. This weak annotation procedure allows for rapid annotation. We only annotate vehicles, specifically the Cityscapes classes: car, truck and bus. We do not consider motorcycle, bicycle, or pedestrian detection in this work.

\subsection{Implementation Details}

In this section we provide specific details on our configurations of each subsystem in this work.

\subsubsection{Detection}

For our proposed method, we modify YOLOP by reducing the default detection threshold to $0.1$ to increase its recall. We also consider the application of autonomous navigation, and note that very far away vehicles are less important for short-term perception and navigation decisions. Therefore, we add in a bounding box size limit filter, removing candidate detections that are less than 0.08\% of the whole image pixel area. This initial filter has two purposes - first, it removes a number of very small false positives, and second the accuracy of our manual annotation reduces below this size.

\subsubsection{Localization}

The RCSv2 dataset is divided into 49 submaps; we follow the standard practice of limiting our VPR retrieval to the submap~\cite{toft2020long}. We perform visual place recognition using the `performance variant' of Patch-NetVLAD. 

During our ablation studies, we also show results where it is assumed that localization is perfect. To do so, we use the ground truth poses provided in the training set of the RCSv2 dataset. We also use the ground truth poses during training, to ensure that the network is trained with the correct corresponding mapping image for each query training image.

\subsubsection{Ablation Study Configurations}

To accurately determine the ideal classifier configuration, we trained several different networks with different designs and also show results with simpler classification architectures. Our ablation systems are listed below:
\begin{itemize}
    \item \textit{Mean L2 distance} between bounding box features
    \item Trained classifier network on GeM pooled \textit{disparity features}
    \item Trained classifier network using \textit{query-only features}
\end{itemize}
where disparity features are the subtraction of the pooled query and mapping bounding box embeddings (instead of concatenation). The L2 distance is the Euclidean difference of the place recognition neural network features that are located within a bounding box.

\subsubsection{Training Parameters}

When training all our classifier network configurations, we used a learning rate of 0.0005 with the Adam optimizer and dropout was set to 0.25. We trained our network with early stopping criterion on the validation F1 score (at detection threshold, $\tau = 0.25$), which stopped training when no improvement is observed for two epochs in a row. We use the model corresponding to the epoch with the highest validation F1 score.

\subsubsection{Evaluation Metrics}

In this work, our focus is on whether or not a given vehicle is detected or not. Our objective is to maximise the number of correct vehicle detections, while minimising the number of false detections. Therefore, we use precision and recall as our metrics, for a given detection threshold. We show precision-recall curves of each classifier, sweeping across different detection thresholds. We also show the system operating point, that is, the precision and recall at the detection decision threshold $\tau$, which is set to $0.25$ for all our experiments, and is also the same detection threshold as YOLOP with default settings. We summarize the PR curves with four summary statistics: F1 score at the operating point, AUC, maximum F1 score and precision at 95\% recall.

\section{Results}\label{sec:results}

In this section, we show the performance of our proposed map matching object detection system on multiple traverses of Oxford RobotCar along with Dark Zurich, and also discuss different ablation studies.

\subsection{Performance of Map Matching Object Detection on Oxford RobotCar}

\begin{figure*}[t!]  %
    \centering
    \includegraphics[width=\linewidth, trim=0cm 0cm 0cm 0cm,clip]{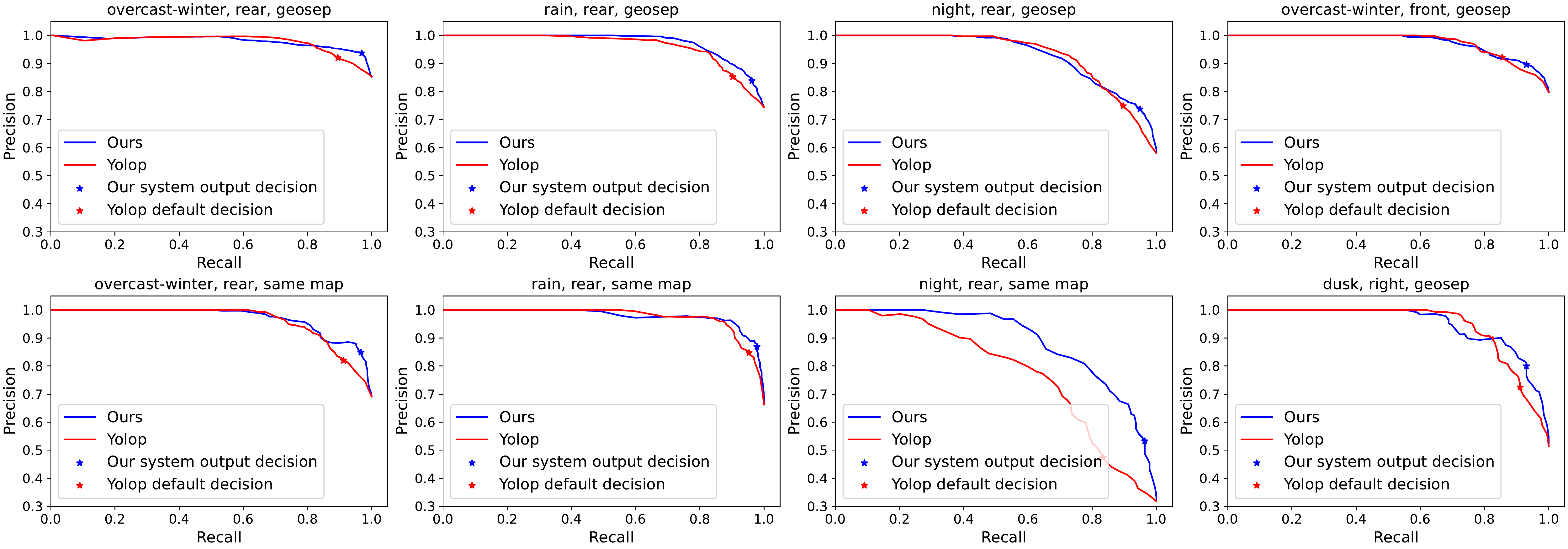}
    \caption{Precision recall curves for eight traverses of Oxford RobotCar. \emph{Detailed legend:} Geosep means that the classifier network has not been trained on the deployment environment. Same map means that the classifier network has seen the deployment environment. The red star is the performance of YOLOP out of the box. The blue star is the performance of the new map-matching addition, building upon the performance of the red star.}
    \label{fig:prcurves}
    \vspace*{-0.2cm}
\end{figure*}

In Figure~\ref{fig:prcurves}, we show the PR curves of our classifier system, which combines the detection scores of both YOLOP and our map matching classifier network. We compare against vanilla YOLOP. We mark the system operating point as a star on each PR curve ($\tau = 0.25$ for both methods). We show results on eight different traverses through the Oxford RobotCar Seasons dataset, across different conditions, cameras and routes (where \emph{geosep} means that the reference map and query images are completely unseen).

As expected, we observe that our approach has a consistent improvement in the precision at high recall, for all eight traverses. This shows that our system is able to remove false positives while still detecting almost every vehicle in each image. This is especially important for autonomous vehicle applications where missing any vehicle detections could have severe consequences. We note that the improvement with our approach is especially large for the traverse \emph{night, rear, same map}. This is a challenging condition for YOLOP by itself, and the map matching network can leverage the mapping images to compensate for these challenging conditions. When the train-test generalization gap increases for \emph{night, rear, geosep}, this improvement reduces in magnitude but still has value.

In Table~\ref{tab:main}, we provide summary statistics for these traverses, including the F1 Score at the operating point (marked with stars in Fig~\ref{fig:prcurves}). Our system has a consistent improvement over YOLOP for all traverses and all metrics. The precision at 95\% recall has the most clear improvement, with the average precision (across all traverses) increasing from 73.3\% to 81.1\%. The greatest improvement is in the traverse \emph{night, rear, same map}, while the smallest improvement is for \emph{overcast-winter, front, geosep}. Since the training data only consisted of images from the rear camera, a larger generalization gap exists for this traverse where front camera images are used instead. While we used rear camera to align with the standard RCSv2 benchmark, front camera results can be expected to improve with front camera training, based on the performance patterns observed for the rear camera.

\begin{table}
  \footnotesize
  \setlength\tabcolsep{0.05cm}

      \caption{Detection Performance using different metrics and dataset traverses.}
    \centering
   \begin{adjustbox}{width=0.5\textwidth}
    \begin{tabular}{lcccccccc}
    \toprule
    & \multicolumn{2}{c}{\textbf{F1 Score}} & \multicolumn{2}{c}{\textbf{AUC}} & \multicolumn{2}{c}{\textbf{Max F1 Score}} & \multicolumn{2}{c}{\textbf{P@95\%R}} \\
    \cmidrule{2-9}
    \textbf{Condition,Cam,Geo} &
    YOLOP &
    Ours &
    YOLOP &
    Ours &
    YOLOP &
    Ours &
    YOLOP &
    Ours \\
    \midrule
    Winter, Rear, Geosep & 
    90.7 &
    \textbf{95.3} &
    97.7 &
    \textbf{98.0} &
    92.2 &
    \textbf{95.3} &
    89.0 &
    \textbf{94.2} \\
    
    Winter, Rear, Same map & 
    86.3 &
    \textbf{90.4} &
    96.1 &
    \textbf{97.0} &
    87.3 &
    \textbf{91.4} &
    77.6 &
    \textbf{88.0} \\
    
    Winter, Front, Geosep & 
    88.8 &
    \textbf{91.3} & 
    97.4 &
    \textbf{97.5} & 
    90.6 &
    \textbf{91.8} &
    85.9 &
    \textbf{88.6} \\

    Rain, Rear, Geosep & 
    87.7 &
    \textbf{89.6} & 
    96.3 &
    \textbf{97.5} & 
    88.0 &
    \textbf{90.4} &
    80.0 &
    \textbf{85.6}\\

    Rain, Rear, Same map & 
    89.7 &
    \textbf{92.0} &
    97.6 & 
    \textbf{97.8} &
    91.6 &
    \textbf{93.1} &
    85.4 &
    \textbf{89.8} \\

    Night, Rear, Geosep & 
    81.6 &
    \textbf{83.0} & 
    93.1 &
    \textbf{93.2} & 
    82.8 &
    \textbf{83.5} &
    68.1 &
    \textbf{73.5} \\

    Night, Rear, Same map & 
    60.4 &
    \textbf{68.6} & 
    77.8 &
    \textbf{88.9} & 
    70.9 &
    \textbf{79.2} &
    36.4 &
    \textbf{55.8} \\

    Dusk, Right, Geosep & 
    80.7 &
    \textbf{86.0} &
    94.4 &
    \textbf{95.0} & 
    86.3 &
    \textbf{87.5} &
    64.1 &
    \textbf{73.0} \\

    \midrule

    Averaged results &
    83.2 &
    \textbf{87.0} &
    93.8 & 
    \textbf{95.6} &
    86.2 &
    \textbf{89.0} &
    73.3 &
    \textbf{81.1} \\
    
    \bottomrule

    \end{tabular}
   \end{adjustbox}
    \label{tab:main}
\end{table}\label{table:results}

\subsection{Performance on Dark Zurich}

To further demonstrate the generalization capabilities of our approach, we tested our system on the Dark Zurich dataset. We show summary statistics in Table~\ref{tab:darkzurich}. 
Despite different camera types and viewing angle in Dark Zurich as compared to the Oxford Robotcar dataset, we can observe that our method is still effective at the default operating point (the F1 Score) and at higher levels of recall, with the precision at $95\%$ recall increasing from $70.8\%$ with YOLOP to $74.5\%$ when we utilize the prior map. We argue that despite a lower overall AUC, for an autonomous vehicle to operate safely without missing detections, a higher value of P@95\% recall is more crucial. 

\begin{table}
  \footnotesize
  \setlength\tabcolsep{0.05cm}
      \caption{Detection Generalization on the Dark Zurich Dataset}
    \centering
    \begin{tabular}{cccccccc}
    \toprule
    \multicolumn{2}{c}{\textbf{F1 Score}} & \multicolumn{2}{c}{\textbf{AUC}} & \multicolumn{2}{c}{\textbf{Max F1}} & \multicolumn{2}{c}{\textbf{P@95\%R}} \\
    \cmidrule(lr){1-2} 
    \cmidrule(lr){3-4}
    \cmidrule(lr){5-6} 
    \cmidrule(lr){7-8}
    YOLOP &
    Ours &
    YOLOP &
    Ours &
    YOLOP &
    Ours &
    YOLOP &
    Ours \\
    \midrule
    71.8 &
    \textbf{74.3} &
    \textbf{87.5} &
    81.9 &
    \textbf{82.9} &
    \textbf{82.9} &
    70.8 &
    \textbf{74.5} \\
    
    \bottomrule

    \end{tabular}
    \label{tab:darkzurich}
\end{table}\label{table:zurich}

\subsection{Ablation Study - Different Map Comparison Techniques}

We ran an ablation study to validate whether or not having access to the prior map is actually beneficial or not, for object detection. It is known that any classification neural network can be taught to detect vehicles in the query image alone, and therefore a prior mapping image may not be required. This can be best analysed by exactly repeating our training process using a network that only ever sees the query images, that is, it will learn to detect objects (without the map) under adverse conditions specific to Oxford's urban environment. We also show ablations using alternative, simpler map comparison techniques: L2 distance and disparity network. These results are shown in Table~\ref{tab:ablate}. We use the F1 score at the detection decision operating point (marked star in Fig~\ref{fig:prcurves}) as one of our metrics for comparing these different ablations. Our second metric is the precision at $95\%$ recall. 

Observing the averaged results, our proposed approach has a higher F1 score at the operating point ($87.0$) and a higher precision at $95\%$ recall ($81.1\%$) compared to the alternative design choices. The second-best performing method is the query-only network. In most instances, the addition of the prior map data provides an improvement in the precision at $95\%$ recall of up to $5\%$. We note the \emph{dusk, right, geosep} traverse is an outlier, where the query only network has a precision of $80.6\%$ at $95\%$ recall. This shows that there is future scope for further inclusion of side-view cameras in the training of our matcher network.
In the disparity network, after query and map features are subtracted, the original feature information is lost, whereas with concatenation the network can leverage the original features to make a more accurate decision, thus performing better than the disparity network.

\begin{table}
  \footnotesize
  \setlength\tabcolsep{0.05cm}
      \caption{Ablating different method designs for map matching}
    \centering
   \begin{adjustbox}{width=0.5\textwidth}
    \begin{tabular}{lcccccccccc}
    \toprule
    & \multicolumn{5}{c}{\textbf{F1 Score at Operating Point}} & \multicolumn{5}{c}{\textbf{Prec @ 95\% Recall}} \\
    \cmidrule{2-11}
    \textbf{Condition,Cam,Geo} &
    YoloP &
    Ours &
    L2 &
    QOnly &
    Disp. &
    YoloP &
    Ours &
    L2 &
    QOnly &
    Disp. \\
    \midrule
    Winter, Rear, Geosep & 
    90.7 &
    \textbf{95.3} &
    91.0 &
    94.5 &
    92.3 &
    89.0 &
    \textbf{94.2} &
    87.9 &
    92.8 &
    91.2 \\
    
    Winter, Rear, Same map & 
    86.3 &
    \textbf{90.4} &
    82.5 &
    87.5 &
    88.7 &
    77.6 &
    \textbf{88.0} &
    72.9 &
    83.6 &
    82.5 \\
    
    Winter, Front, Geosep & 
    88.8 &
    \textbf{91.3} & 
    89.6 &
    \textbf{91.3} & 
    86.9 &
    85.9 &
    \textbf{88.6} &
    81.9 &
    88.5 &
    83.9 \\

    Rain, Rear, Geosep & 
    87.7 &
    \textbf{89.6} & 
    86.8 &
    88.1 & 
    87.7 &
    80.0 &
    \textbf{85.6} &
    79.9 & 
    83.8 &
    80.4 \\

    Rain, Rear, Same map & 
    89.7 &
    \textbf{92.0} &
    85.7 & 
    88.8 &
    91.1 &
    85.4 &
    89.8 &
    79.7 &
    \textbf{91.3} &
    88.0 \\

    Night, Rear, Geosep & 
    81.6 &
    \textbf{83.0} & 
    74.3 &
    80.1 & 
    80.0 &
    68.1 &
    \textbf{73.5} &
    64.1 &
    72.4 &
    66.5 \\

    Night, Rear, Same map & 
    60.4 &
    \textbf{68.6} & 
    49.2 &
    62.8 & 
    67.3 &
    36.4 &
    \textbf{55.8} &
    34.5 &
    53.8 &
    49.7 \\

    Dusk, Right, Geosep & 
    80.7 &
    \textbf{86.0} &
    75.3 &
    83.7 & 
    81.4 &
    64.1 &
    73.0 &
    64.1 &
    \textbf{80.6} &
    67.2 \\

    \midrule

    Averaged results &
    83.2 &
    \textbf{87.0} &
    79.0 & 
    84.6 &
    84.4 &
    73.3 &
    \textbf{81.1} &
    70.6 &
    80.9 &
    76.2 \\
    
    \bottomrule

    \end{tabular}
    \end{adjustbox}
    \label{tab:ablate}
\end{table}\label{table:ablation}

\subsection{Ablation Study - with Ground Truth Localization}

In this study, we replace the visual place recognition localization with the six degree of freedom ground truth (GT) poses provided by the RobotCar Seasons v2 dataset. This ensures that the detection performance is not reduced due to incorrect retrieval by VPR. We show the vehicle detection performance \textit{with VPR} versus \textit{with GT} in Table~\ref{tab:gt}. We use the \emph{same map} setting for this study. We observe little difference in the detection performance, with a slight improvement across all metrics when ground truth poses are used. Interestingly, in some cases, VPR coincidentally performs better than GT, where the detection region of an incorrect reference image happened to match more with the query region than deeming it as false positive. Overall, this study shows that using a VPR system instead of requiring ground truth localization is a viable solution for map-based vehicle detection.

\begin{table}
  \footnotesize
  \setlength\tabcolsep{0.15cm}
    \caption{Detection performance with VPR localization versus ground truth localization for rear camera and same map setting.}
    \centering
   \begin{adjustbox}{width=0.45\textwidth}
    \begin{tabular}{lcccccccc}
    \toprule
    & \multicolumn{2}{c}{\textbf{F1 Score}} & \multicolumn{2}{c}{\textbf{AUC}} & \multicolumn{2}{c}{\textbf{Max F1}} & \multicolumn{2}{c}{\textbf{P@95\%R}} \\
    \cmidrule{2-9}
    \textbf{Condition} &
    VPR &
    GT &
    VPR &
    GT &
    VPR &
    GT &
    VPR &
    GT \\
    \midrule
    Winter & 
    \textbf{90.4} &
    90.3 &
    97.0 &
    \textbf{97.1} &
    91.4 &
    \textbf{91.7} &
    \textbf{88.0} &
    87.4 \\

    Rain& 
    92.0 &
    \textbf{92.2} &
    97.8 & 
    \textbf{97.9} &
    93.1 &
    \textbf{93.5} &
    89.8 &
    \textbf{90.0} \\

    Night& 
    68.6 &
    \textbf{69.8} & 
    88.9 &
    \textbf{89.2} & 
    \textbf{79.2} &
    78.9 &
    55.8 &
    \textbf{58.2} \\
    
    \bottomrule

    \end{tabular}
   \end{adjustbox}
    \label{tab:gt}
\end{table}\label{table:gtresults}

\begin{figure*}  %
    \centering
    \includegraphics[width=\textwidth, trim=0cm 20cm 0cm 0cm,clip]{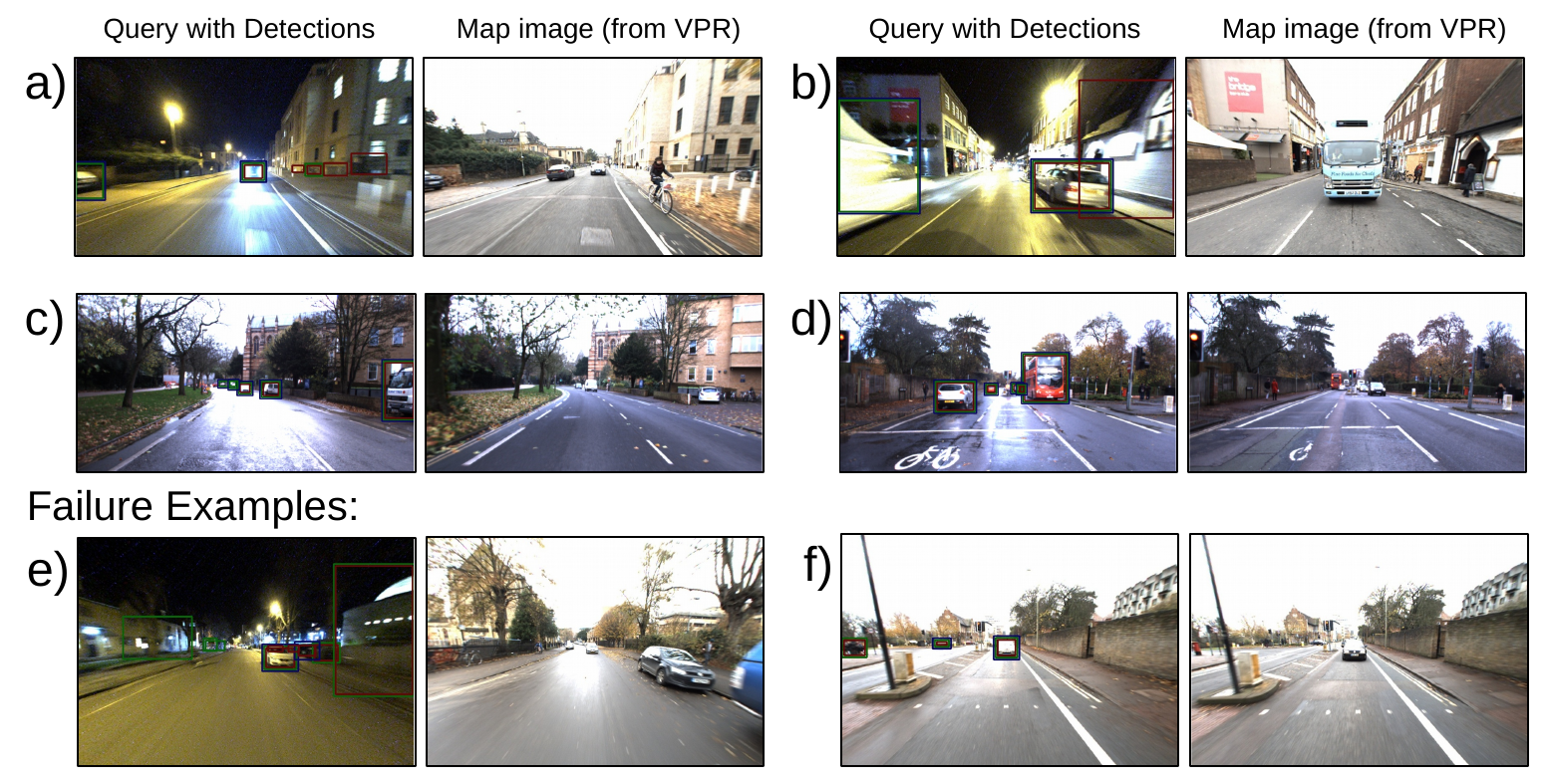}
    \caption{We show six example image pairs from the Oxford Robotcar dataset, displaying the operation of our proposed map aided object detector. In each image pair, the left image shows the query image with detection bounding boxes and the right image shows the map image used for the current query. A blue box denotes a ground truth detection, a red box denotes a detection from YOLOP (at the standard operating point) and a green box denotes a detection from our approach. We show four success examples and two failure cases.}
    \label{fig:example}
\end{figure*}
\subsection{Qualitative Results}

In Figure~\ref{fig:example}, we show six example image pairs of our system in action. In example A, YOLOP experiences several false positives, while our detector has fewer false positives and also detects a vehicle that YOLOP missed. In example B, YOLOP misses a vehicle with severe motion blur and also has a false positive detection. In example C, two far away vehicles are missed by YOLOP but detected with our method. In example D, both systems are equal in performance. Example E shows a failure case, where localization has failed and the wrong map image is being used. As expected, our approach performs poorly when this occurs. In the final example (F), both YOLOP and our map matching classifier incorrectly classify a group of pedestrians as a vehicle; note, the pedestrians are also technically a map difference and therefore this situation is a natural byproduct of training to detect map differences.

\section{Conclusion}

A key takeaway from this work is that dynamic object detection -- the detection of objects that are move-able -- can be considered not just as a task based on live observation, but one with location-grounding, where the observed scene differs from the previously observed scene in the prior reference map. Essentially, the prior map can be seen as a prior that specifies locations where dynamic objects are most likely to be located. This then leads to the future work question as to whether an object detector can be trained in entirety with inputs from both the current query and the matching map image, for improved recognition of dynamic objects. 

In this work, we demonstrate and provide a method for improving the ability to detect vehicles in an autonomous driving scenario, above the performance of an existing state-of-the-art vehicle detector. 
Our two-stage approach is an add-on to an existing object detector, which uses an initial large set of detection candidates and then refines these candidates based on a confidence score that utilises the prior map. In summary, we show that localization can be utilized to boost the performance of dynamic object detection, and we experimentally demonstrate using an autonomous driving scenario on the public datasets: Oxford RobotCar and Dark Zurich.

\bibliographystyle{IEEEtran}
\bibliography{references,sg}

\end{document}